\documentclass[sigconf]{aamas}

\usepackage{balance} 

\setcopyright{ifaamas}
\acmConference[AAMAS '25]{Proc.\@ of the 24th International Conference
on Autonomous Agents and Multiagent Systems (AAMAS 2025)}{May 19 -- 23, 2025}
{Detroit, Michigan, USA}{A.~El~Fallah~Seghrouchni, Y.~Vorobeychik, S.~Das, A.~Nowe (eds.)}
\copyrightyear{2025}
\acmYear{2025}
\acmDOI{}
\acmPrice{}
\acmISBN{}

\acmSubmissionID{1294}

\title[AAMAS-2025 Formatting Instructions]{CSAOT: Cooperative Multi-Agent System for Active Object Tracking}

\author{Hy Nguyen*}
\affiliation{
  \institution{Applied Artificial Intelligence Institute, Deakin University}
  \city{Melbourne}
  \country{Australia}}
\email{hy.nguyen@deakin.edu.au}

\author{Bao Pham*}
\affiliation{
  \institution{Applied Artificial Intelligence Institute, Deakin University}
  \city{Melbourne}
  \country{Australia}}
\email{phambao.atwork@gmail.com}

\author{Hung Du}
\affiliation{
  \institution{Applied Artificial Intelligence Institute, Deakin University}
  \city{Melbourne}
  \country{Australia}}
\email{hung.du@deakin.edu.au}

\author{Srikanth Thudumu}
\affiliation{
  \institution{Applied Artificial Intelligence Institute, Deakin University}
  \city{Melbourne}
  \country{Australia}}
\email{srikanth.thudumu@deakin.edu.au}

\author{Rajesh Vasa}
\affiliation{
  \institution{Applied Artificial Intelligence Institute, Deakin University}
  \city{Melbourne}
  \country{Australia}}
\email{rajesh.vasa@deakin.edu.au}

\author{Kon Mouzakis}
\affiliation{
  \institution{Applied Artificial Intelligence Institute, Deakin University}
  \city{Melbourne}
  \country{Australia}}
\email{kon.mouzakis@deakin.edu.au}

\begin{abstract}

Object Tracking is essential for many computer vision applications, such as autonomous navigation, surveillance, and robotics. Unlike Passive Object Tracking (POT), which relies on static camera viewpoints to detect and track objects across consecutive frames, Active Object Tracking (AOT) requires a controller agent to actively adjust its viewpoint to maintain visual contact with a moving target in complex environments. Existing AOT solutions are predominantly single-agent-based, which struggle in dynamic and complex scenarios due to limited information gathering and processing capabilities, often resulting in suboptimal decision-making. Alleviating these limitations necessitates the development of a multi-agent system where different agents perform distinct roles and collaborate to enhance learning and robustness in dynamic and complex environments. Although some multi-agent approaches exist for AOT, they typically rely on external auxiliary agents, which require additional devices, making them costly. In contrast, we introduce the Collaborative System for Active Object Tracking (CSAOT), a method that leverages multi-agent deep reinforcement learning (MADRL) and a Mixture of Experts (MoE) framework to enable multiple agents to operate on a single device, thereby improving tracking performance and reducing costs. Our approach enhances robustness against occlusions and rapid motion while optimizing camera movements to extend tracking duration. We validated the effectiveness of CSAOT on various interactive maps with dynamic and stationary obstacles.
\end{abstract}


\begin{document}


\pagestyle{fancy}
\fancyhead{}


\maketitle 


\section{Introduction}

Object tracking is a fundamental task in computer vision, broadly categorized into passive and active tracking ~\cite{DBLP:journals/corr/abs-2201-13066}. Passive Object Tracking (POT) identifies a target object from an input video sequence recorded by a static camera. POT operates without influencing or interacting with the environment or the target object; its sole task is to detect the target object across consecutive video frames. In contrast, Active Object Tracking (AOT) involves an interactive agent that actively adjusts its viewpoint to maintain continuous visual contact with the target object as it moves through the environment. This makes AOT inherently more dynamic than POT, requiring real-time decision-making capabilities to adapt to rapid and often unpredictable changes in the target's movement, environmental conditions, and potential obstacles. 

Deep Reinforcement Learning (DRL) is a subset of machine learning that combines reinforcement learning with deep learning techniques to enable agents to make decisions in complex environments. DRL's strengths lie in its ability to learn optimal policies through trial and error, leveraging large amounts of data and powerful neural networks to approximate value functions ~\cite{nguyen2023uav}. This adaptability allows DRL to handle high-dimensional state spaces and learn from delayed rewards, making it particularly effective for tasks requiring continuous interaction and decision-making. In the context of Active Object Tracking (AOT), DRL is the most suitable method because it enables agents to dynamically adjust their tracking strategies in real-time, responding to the unpredictable movements of target objects while optimizing tracking accuracy and efficiency.

Existing DRL solutions to AOT primarily rely on a single-agent approach, where one agent performs all actions for the tracking task (e.g., detection, navigation, obstacle avoidance) \cite{jiang2019multi}. While these solutions can be effective in static environments, they often limit the agent's ability to adapt in dynamic environments~\cite{padakandla2021survey}. This can be due to challenges, such as occlusions, varying speeds, and uncertain changes in the target object's trajectory. To alleviate these challenges, distributing actions into the multi-agent system where each agent performs a single action can be a potential solution \cite{jiang2019multi}. In addition, Multi-Agent Deep Reinforcement Learning (MADRL) can be employed to facilitate agent interaction for AOT in dynamic environments ~\cite{padakandla2021survey}. In MADRL, agents communicate, coordinate and learn from their experiences, improving their decision-making processes and adaptability to the unpredictable behavior of target objects ~\cite{Silver2016MasteringTG, Mnih2015HumanlevelCT}. Note that, there exists both individual goals and collective goals in MADRL. This leads to the challenge of designing a reward function that support the optimal performance of the entire system ~\cite{wang2023synergistic}. Current MADRL methods for AOT mainly focus on designing auxiliary agents that operate externally to the main agent. In this setup, the master agent is responsible for the tracking task, while the auxiliary agents gather and transmit supplementary information to enhance the master agent's decision-making capabilities ~\cite{ning2024survey}. Although this approach can improve tracking accuracy, it necessitates the use of additional devices and resources, leading to increased costs and complexity in system deployment. 

To address these challenges, we propose a role-based approach that enables multiple agents to collaborate within a single device. This innovative design not only reduces costs associated with additional hardware but also minimizes communication overhead. Additionally, we propose a novel mechanism called Mixture of Policy (MoP) to learn a policy for each agent. The MoP employs a gating mechanism to coordinate multiple smaller policy networks, with each policy acting as an expert tailored to handle specific scenarios encountered during the tracking task. This approach significantly reduces inference time, which is crucial given the multiple agents involved in the system, enabling faster decision-making without sacrificing performance. Additionally, the use of expert policies enhances accuracy, as each agent can select the most relevant policy based on the current context. By combining the advantages of the MoP mechanism with a role-based collaboration framework, our method aims to optimize both efficiency and effectiveness in real-time AOT applications. In summary, our contribution to this work includes:
\begin{itemize}
    \item Proposed a novel framework CSAOT for cooperative multi-agent deep reinforcement learning for AOT task on a single device.
    \item Introduced MoP, a novel mechanism to learn policy efficiently.
    \item Adapt the framework to AOT task by applying subtask-based reward functions.
    \item Evaluate the proposed framework on a simulated environment, empirically proving the system's performance.
\end{itemize}


\section{Related work}

\subsection{Object tracking task and existing solutions}
With AOT, its application in real-time environment is broader, hence different approaches have been developed to solve this task. Several studies have leveraged reinforcement learning and multi-camera collaboration to improve active object-tracking performance. Luo et al. \cite{luo2019end} introduced an end-to-end tracking system that integrates object tracking and camera control into a single reinforcement learning framework. The system demonstrates adaptability in real-world scenarios by training in virtual environments and deploying on physical robots. Similarly, Li et al. \cite{li2020pose} tackled object-tracking challenges using pose-assisted multi-camera collaboration. Their approach allows cameras to share positional data, enabling more accurate tracking in environments with multiple obstructions and complex dynamics. These methods highlight the power of collaborative and learning-based strategies to optimize tracking performance across diverse environments.

Another previous sub-field in object tracking focuses on active feature selection and motion detection to enhance the robustness and efficiency of object tracking. Zhang et al. \cite{zhang2013robust} applied active learning techniques to feature selection, ensuring that the most relevant features are used for monitoring, which improves performance in real-time applications. Denzler and Paulus \cite{denzler1994active} proposed a two-stage active vision system that separates motion detection from tracking, enhancing performance in dynamic and unpredictable settings. Different approaches focus on active contour models and particle filters to enhance object tracking. Silva et al. \cite{silva2016object} and Lefèvre and Vincent \cite{lefevre2004real} used active contour models to improve object segmentation and tracking in real-time by adapting to the object's shape and contours.

Researchers have also utilized deep learning and anti-distractor mechanisms to address tracking in complex environments. Xi et al. \cite{xi2021anti} proposed a novel anti-distractor approach specifically designed for 3D environments with multiple distractors, using deep learning to ensure that the target object remains the focus even in the presence of other moving objects. Lei et al. \cite{lei2022active} extended active object tracking to space manipulators, employing deep reinforcement learning to track objects in space, tackling the challenges posed by low-gravity environments. These solutions showcase the flexibility of deep learning in overcoming challenges related to complex environments and distractors.

\subsection{Multi-agent Deep Reinforcement Learning}
Multi-agent deep reinforcement learning (MADRL) combines multi-agent systems and deep reinforcement learning (DRL) to address complex challenges ~\cite{Foerster_Farquhar_Afouras_Nardelli_Whiteson_2018}. It extends traditional reinforcement learning to environments with multiple interacting agents, each learning to optimize its rewards while collaborating or competing with others. MADRL emerged from the need for multiple agents to work together in real-world scenarios with shared or conflicting objectives. Value-based methods were introduced commonly, each with a different approach to diversifying the range of tasks MADRL can solve. Real-world applications include autonomous vehicles avoiding collisions and robotic teams performing joint tasks. Centralized training with decentralized execution (CTDE) is a popular MADRL paradigm ~\cite{zhou2023centralizedtrainingdecentralizedexecution}. Agents are trained centrally with global information and later deployed with partial information. Value decomposition methods aim to decompose the global value function into individual value functions, facilitating credit assignment and enhancing scalability. Notable contributions include QMIX ~\cite{rashid2018qmixmonotonicvaluefunction} and Value Decomposition Networks (VDN) ~\cite{sunehag2017valuedecompositionnetworkscooperativemultiagent}, which have been successful in cooperative games with shared reward signals.

Policy gradient methods in MADRL focus on optimizing action-selection policies. Approaches like Multi-Agent Proximal Policy Optimization (MAPPO) ~\cite{yu2022surprisingeffectivenessppocooperative} and Multi-Agent Deep Deterministic Policy Gradient (MADDPG) ~\cite{lowe2020multiagentactorcriticmixedcooperativecompetitive} handle continuous action spaces, which are crucial for complex environments. These are all Proximal Policy Optimization (PPO) and Trust Region Policy Optimization (TRPO) ~\cite{schulman2017proximalpolicyoptimizationalgorithms} adaptations to ensure stable learning within multi-agent scenarios. Explicit communication is vital for effective collaboration in multi-agent environments, which can be shown through approaches like CommNet ~\cite{sukhbaatar2016learningmultiagentcommunicationbackpropagation} and DIAL ~\cite{foerster2016learningcommunicatedeepmultiagent}, which enable agents to share information efficiently, improving performance in scenarios with partial observability. Challenges in MADRL include scalability, coordination, practical exploration-exploitation trade-offs, and reward design that balances individual and group incentives is also tricky. Current research explores hierarchical reinforcement learning and role-based approaches to address these issues by simplifying coordination.

MADRL has broad applications, such as autonomous driving, cooperative robotics, and finance. Specifically in autonomous driving, agents coordinate to ensure the safety and efficiency of the vehicle. In cooperative robotics, drones or robots collaborate to complete tasks like search and rescue. MADRL has also been used in real-time strategy games for agent coordination. Multi-agent deep reinforcement learning is an evolving area with significant potential for solving complex problems. Despite progress in training methods, challenges like scalability and reward design remain. Continued research is critical to fully realizing the potential of MADRL across various domains.

\subsection{Mixture of Experts}
The Mixture of Experts (MoE) model, a significant advancement in deep learning, has revolutionized the field by significantly improving scalability and efficiency in large-scale tasks. Shazeer et al. \cite{shazeer2017outrageously} introduced the Sparsely-Gated Mixture of Experts layer. This key innovation addresses the issue of high computational costs by activating only a subset of experts for each input. This not only reduces the computational burden but also maintains or improves performance. Particularly in tasks such as machine translation and language modeling, this approach enhances the efficiency of training large-scale models, making them feasible to deploy in practice. The MoE model has become a fundamental component in modern large-scale neural networks, where computational resource management is critical.

The Mixture of Experts framework has proven particularly valuable in multi-task learning, enabling the modeling of relationships between tasks while preserving task-specific performance. Ma et al. \cite{ma2018modeling} developed a Multi-Gate Mixture of Experts model, which employs multiple gating networks to assign different experts to different tasks dynamically. This approach improves shared learning and task-specific specialization, essential in multi-task scenarios where other tasks may have distinct requirements. The model's ability to selectively share knowledge between functions while maintaining the independence of tasks that require unique representations underscores the versatility of the MoE framework in handling diverse and competing learning objectives in a single unified model.

\section{Preliminaries}
\subsection{Problem formulation}
Multi-agent cooperative tasks with decentralized execution can be represented under the original Dec-POMDP, which consists of a tuple \begin{equation}
G = \langle S, A, P, O, \Omega, R, C, n, \gamma \rangle,
\end{equation}
Here, \( C \) is a set of \( n \) agents, and \( s \in S \) is the state gathered from the environment. Every agent \( i \) retrieves its observation \( o_i \in \Omega \) drawn from the observation function \( O(s, i) \), which helps decide the next action \( a_i \in A \), creating a joint action \( a \in A^n \), which leads to the next state \( s' \) based on a transition function \( P(s'|s, a) \), and observing a reward \( r = R(s, a) \) which depends on the discount factor \( \gamma \in [0, 1] \). Each agent has local action-observation history \( \tau_i \in \mathcal{T} \equiv (\Omega \times A)^* \). To simplify the problem, we decided to set the discount factor  \( \gamma = 1 \)

\subsection{Task formulation}
Along with the general Dec-POMDP problem formulation, some unique properties come from the nature of the task:
\begin{itemize}
    \item \textbf{Unified observation}: Unlike other cooperative or competitive MADRL systems, all agents $\pi_i$ in our system operate on the same device, receiving their observations $o_i$ from a common source, which includes the current frame $f_t$, acceleration $\alpha_t$, speed $s_t$, and steering angle $\delta_t$.
    \item \textbf{System action format}: The action space of the system's final navigation agent includes a tuple \((x, y)\), where \(x\) is the acceleration rate (\(x \in [-0.5, 0.5]\), negative values mean going in the reverse way) measured in \(m / s^2\) and \(y\) is the turning angle (\(y \in [-0.5, 0.5]\), negative values denote left turns and positive values denote right turns) measured in radians. The action spaces of other auxiliary agents are different based on their assigned tasks, which is explained in Section \ref{sec:CSAOT}.
    \item \textbf{Fully decentralized system}: All agents in the system work in their own unique action space. Under this assumption, collaborative multi-agent reinforcement learning algorithms such as MAPPO don't have the exact nature of our defined task. Hence, we adapted the Decentralized Training - Decentralized Execution (DTDE) approach for our solution.
    \item \textbf{Continuous action space}: For this task, all features collected from observation are scalable actions, such as changing the bounding box, acceleration, and camera angle. As the number of available action families increases, the action space rises remarkably, significantly increasing the task's complexity. 
\end{itemize}
\subsection{Mixture of Experts}
MoE (Mixture of Experts) has different variants, differentiated by the gating mechanism. We consider \(\text{Top-}K \) MoE in this framework due to its balance between computational cost and accuracy. Let $\mathbf{x} \in \mathbb{R}^d$ be the input, and let $f_i(\mathbf{x}; \theta_i)$ represent the expert networks. The top $K$ experts are selected based on their gating probabilities:
\begin{equation}
S_K = \text{Top}K \left(g(\mathbf{x}; \phi)\right) = \{i_1, i_2, \dots, i_k\}.
\end{equation}
The remaining elements in \(S_K\) are reweighted for the corresponding expert's impact. The output of MoE is the weighted sum of the selected experts:
\begin{equation}
y(\mathbf{x}) = \sum_{i \in S_K} \frac{p_i}{\sum_{j \in S_K} p_j} \cdot f_i(\mathbf{x}; \theta_i).
\end{equation}
Ultimately, this hybrid mechanism allows the system to adapt to more scenarios, improving its robustness.
\subsection{Proximal Policy Optimization}
Proximal Policy Optimization (PPO) ~\cite{schulman2017proximalpolicyoptimizationalgorithms} is a reinforcement learning algorithm designed to improve traditional policy gradient methods by ensuring more stable and reliable training. We prefer this method over other reinforcement algorithms because of its decent sample efficiency. PPO ensures that considerable updates to the policy are controlled; hence, the policy is stable during the training process. This is achieved through a clipping mechanism that limits the changes in the policy probability ratio. The key idea is to maximize the clipped objective:

\begin{equation}
L^{CLIP}(\theta) = \mathbb{E}_t \left[ \min \left( r_t(\theta) \hat{A}_t, \text{clip}\left( r_t(\theta), 1 - \epsilon, 1 + \epsilon \right) \hat{A}_t \right) \right]
\end{equation}

where $r_t(\theta)$ is the probability ratio between the new policy and the old policy:

\begin{equation}
r_t(\theta) = \frac{\pi_\theta(a_t | s_t)}{\pi_{\theta_{\text{old}}}(a_t | s_t)}
\end{equation}

and $\hat{A}_t$ is the advantage function at timestep $t$, which helps guide the optimization towards better actions. The clipping parameter $\epsilon$ controls how far the new policy deviates from the old policy. Another advantage of using PPO is that it can adapt to continuous action space, which makes it fit our specific task.
\section{CSAOT Framework} \label{sec:CSAOT}
\subsection{Overall of the framework}
\begin{figure*}[h]
    \centering
    \includegraphics[width=0.9\textwidth]{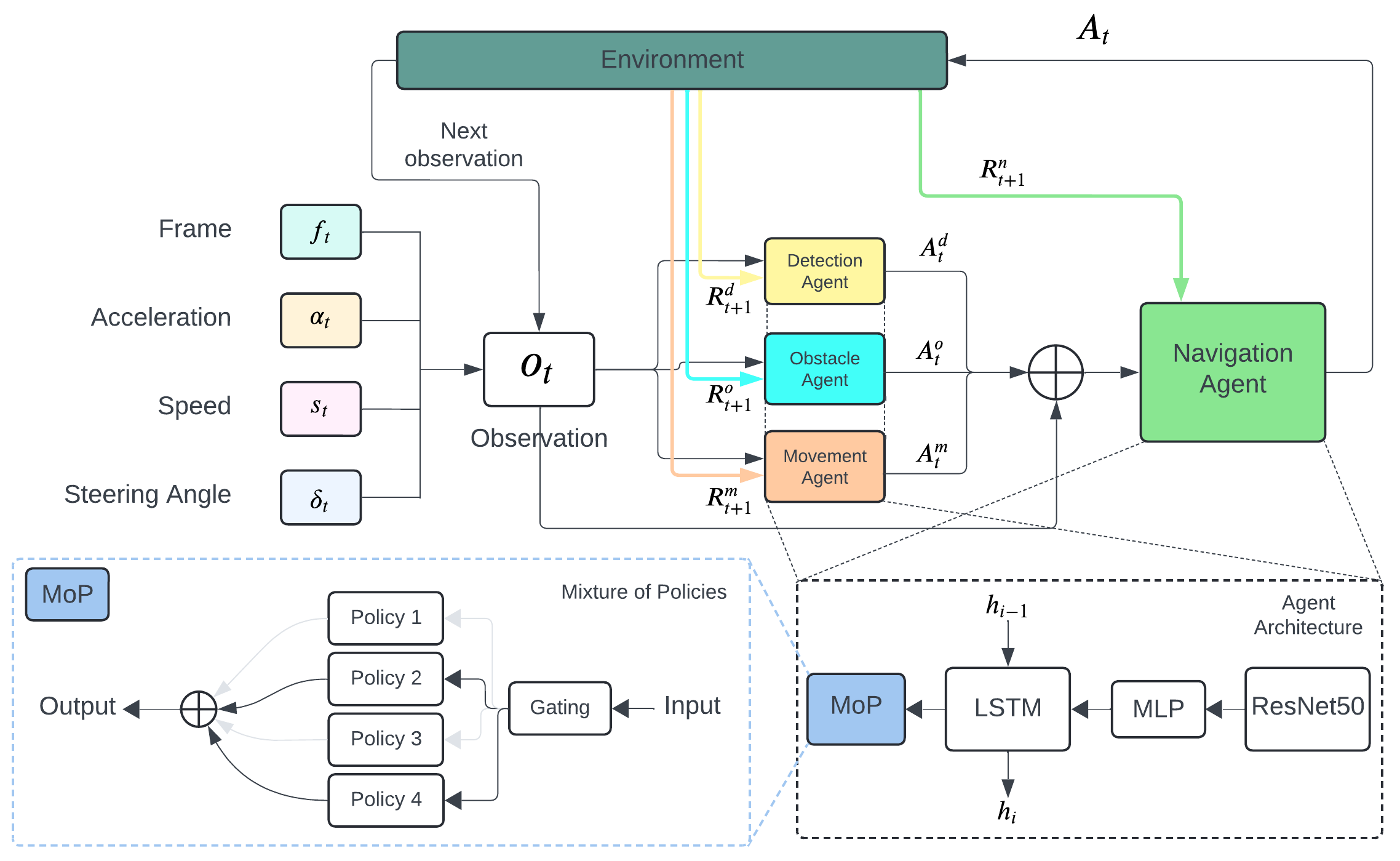}
    \caption{CSAOT framework: Each agent (Detection Agent, Obstacle Agent, Movement Agent) is responsible for different tasks and operates on the same device, using observations from the same source. The final navigation agent takes the outputs of these three agents, along with the original observation, as its input and produces the navigation actions for the tracker. Each agent is implemented using a Mixture of Policy (MoP) mechanism.}
    \label{fig:architecture}
\end{figure*}
\subsubsection{Framework architecture}
Figure \ref{fig:architecture} shows the overall architecture of our proposed method. In detail, we designed the system as a two-layer hierarchical system, with the first layer extracting the core information based on the subtasks from the ultimate objective of tracking the moving target (detection, obstacle, movement) and the second layer used for the final decision-making phase. The raw current frame is processed by a ResNet50 ~\cite{he2015deepresiduallearningimage} pre-trained model for feature extraction. The subtasks we mentioned earlier include:

\begin{itemize}
\item \textbf{Object detection}: This agent aims to predict the object's bounding box in the frame. Specifically, its action space comprises 4 values \(a_d = (x_l, y_l, x_r, y_r)\), denoting the top-left and bottom-right corner of the predicted bounding box. This gives the decision-making agent the target's position on the frame, which is very helpful in choosing the next action.
\item \textbf{Object movement}: This agent predicts the center of the target within the frame, which the format is \(a_n = (x_c, y_c)\). Although this can be inferred from the bounding box given by the previous agent, we hypothesize that the center provided by the bounding box relies heavily on the slight change of any corner. This architecture helps ensure the robustness of the information coming out from this first layer for reasonable action consideration. 
\item \textbf{Obstacle avoidance}: Unlike the previous agents, this agent considers the surroundings as its central focus. The action space of this agent is \(a_a = (x)\), where \(x\) shows the distance of the nearest obstacle existing in the frame. This detection is crucial to the tracker's movement since it's prohibited from collision with any obstacle.
\item \textbf{Action decision}: This agent lies in the second layer, which combines the encoded image and all the predictions from the first layer as the input. The action \(a_i = (\Delta v, \Delta \alpha)\) is used as the final action of the system. This consists of \(\Delta v\) which is the acceleration rate, and \(\Delta \alpha\) denotes the steering angle.
\end{itemize}

Each agent processes two main phases: information encoding and retrieving action. The information encoding phase includes an MLP (Multi-Layer Perceptron) to extract information from the image and a memorial module to combine the current and previous states. Input feature \(e_o\) is encoded to \(o_i\) through the MLP block, which becomes the input for the corresponding LSTM block to form the final encoded state \(e_i\), based on the long-term information from the previous states \(h_{i-1}\): 
\begin{equation} e_o = MLP(o_i),\end{equation}
\begin{equation} h_i, e_i = LSTM(h_{i-1}, e_o)\end{equation}
For sequential tasks like AOT, a memorial module for each agent is needed to learn about the environment efficiently through the last step observations. Previously, GRU was the most common choice for this specific task, although it's been outperformed by other sequential models due to its computational efficiency. In this case, LSTM is chosen for its performance in memorizing historical data ~\cite{chung2014empiricalevaluationgatedrecurrent}. 

The retrieving action phase is the second phase, which takes the encoded information from the previous phase as the input for the policy network. In detail, given the previous phase's output \(e_i\), the policy network computes the final action by
\begin{equation}
a_i = MoP(e_i)
\end{equation}
where \(MoP\) is our proposed mechanism for calculating the current policy, introduced in Section \ref{sec:MoP}.
\subsubsection{Reward construction}
The reward construction contains multiple component rewards for each agent. The final layer agent would take the global reward as its main reward. The target's actual bounding box in the frame \(m = \{(x_i, y_i)|i \in [1, n]\}\) needs to be calculated to calculate the rewards. For states where the tracker collides with any other object in the environment, its reward for that state is -50, and the corresponding episode ends immediately. In every reward component, we designed a proper scaling \(\lambda\) to appropriately adjust the task's priority. The reward can be decomposed into these smaller parts:
\begin{itemize}
    \item \textbf{Tracking reward}: \(R_{track}\) takes the area of the target's bounding box compared to the frame's area. The motivation is to ensure the tracker keeps an appropriate distance from the object. Specifically, this can be written as: 
    \begin{equation}
    R_{track} = min(\dfrac{S_{m_i}}{S_{frame} * 0.25}, 2 - \dfrac{S_{m_i}}{S_{frame} * 0.25}) \cdot \lambda_{track}
    \end{equation}
    where \(S\) denotes the area of a given polygon.
    \item \textbf{Navigation reward}: \(R_{nav}\) is calculated based on the center of \(m\) using the following function:
    \begin{equation}
    center(box) = (\dfrac{1}{n} \cdot \sum_{i}x_i, \dfrac{1}{n} \cdot \sum_{i}y_i)
    \end{equation}
    After getting the center of the corresponding bounding box, the Manhattan distance is used to calculate the reward\(R_{nav}\) can be written as:
    \begin{equation}
    R_{nav} = -\dfrac{Manhattan(center(m), center(frame))}{Manhattan((0, 0), center(frame))} \cdot \lambda_{nav}
    \end{equation}
    \item \textbf{Behavioural reward}: This is treated as a behavioral guide for the tracker, which helps maintain consistency in the movement during the tracking process. It includes the following components:
    \begin{equation}
    R_{move} = \begin{cases}
  -3, & \text{if } \text{speed} = 0 \text{ and } \Delta v_i \leq 0 \\
  0,  & \text{otherwise} \end{cases}\end{equation}
  \begin{equation}R_{steer} = \begin{cases}
  -3, & \text{if } \text{speed} = 0 \text{ and } \Delta \alpha_i \neq 0 \\
  0,  & \text{otherwise} \end{cases}\end{equation}
  \begin{equation}
  R_{diff} = |\Delta v_i - \Delta v_{i-1}| \cdot \lambda_{diff}
    \end{equation}
    where \(R_{move}\) penalized the tracker whenever it brakes while not running, \(R_{steer}\) penalized the tracker when it performs steer when the vehicle stops, and \(R_{diff}\) penalized the tracker for any sudden change in the acceleration.
\end{itemize}
\subsection{Novel adaptations}
\subsubsection{Task-based component rewards}
Besides the global reward for the last layer agent, we designed unique reward functions for each agent in the first layer. This helps the agents' learning speed stay stable within the training process:
\begin{itemize}
    \item \textbf{Detection Agent}: For the bounding box prediction, one direct way to assess the correctness of the output is by using the Intersection over Union (IoU) metric. Specifically, the reward is computed as:
    \begin{equation}
    R_{detect} = IoU(m, a_d) \cdot \lambda_{detect}
    \end{equation}
    \item \textbf{Obstacle Agent}: The obstacle distance to the tracker can be assessed by the absolute distance between the actual distance (Calculated through LIDAR sensors information) and the predicted distance
    \begin{equation}
    R_{obstacle} = |a_a - d_{true}| \cdot \lambda_{obstacle}
    \end{equation}
    \item \textbf{Movement Agent}: With tracking the center of the target, one standard reward that can be utilized is the Manhattan distance between the ground truth and the prediction
    \begin{equation}
    R_{movement} = \text{Manhattan}(a_n, center(m)) \cdot \lambda_{movement}
    \end{equation}
\end{itemize}
As the system learns through a single source of observation, learning only through the global reward could cause a bottleneck in the learning speed between different agents in the system. Furthermore, the global reward doesn't seem relevant to the first layer's agents; hence, learning those could be less efficient than learning those in the second layer. For these reasons, we designed rewards that directly rely on their actions. 
\subsubsection{Mixture of Policies} \label{sec:MoP}
In such a complex task like ASOT with continuous action space, it's harder for the system to explore its action space to a sufficient level, which leads to poor generalization to different situations that the environment may provide. Motivated by that critical weakness, an alternative mechanism to help improve this is needed. The Mixture of Experts technique is widely known for its scalability, low inference cost, and adversarial robustness. With all those properties that MoE has, we chose to integrate MoE into the policy network based on the hypothesis that this would help the agent learn a diverse range of policies based on current state observations, forming our proposed method Mixture of Policies (MoP) method.  Formally, MoP can be written as:
\begin{equation}
MoP(\mathbf{o, K}) = \sum_{i \in S_K} \frac{w_i}{\sum_{j \in S_K} w_j} \cdot p_i(\mathbf{o}; \theta_i).
\end{equation}
Here, \(K\) is the number of chosen experts to retrieve the final action, and \(w_i\) is the weight for the corresponding sub-policy network \(p_i\). \(p = {p_1, p_2, ..., p_n}\) are the individual policy networks, which are dynamically chosen in specific cases to participate in the policy combination. The way \(w\) is calculated in this architecture is the same as the previously shown MoE method. 
 
\section{Experiments}
For all scenarios in our experiment, the hyperparameter values are shown in table \ref{tab:hyperparameter}.
\begin{table}[h!]
    \centering
    \begin{tabular}{|c|c|}
        \hline
        \textbf{Hyperparameter} & \textbf{Value} \\
        \hline
        \textit{Training GPU} & NVIDIA RTX 4060 6GB \\
        \hline
        \textit{Training CPU} & Intel(R) Core(TM) i5-12400F\\        
        \hline
        \textit{Training RAM} & 32GB\\        
        \hline
        \textit{Epoch per sample} & 2\\
        \hline
        \textit{Number of expert networks (MoP)} & 4\\
        \hline
        \textit{Number of chosen networks \(K\) (MoP)} & 2\\
        \hline
        \textit{Learning rate} & 0.003\\
        \hline
        \textit{\(\epsilon\)} & 0.99\\
        \hline
        \textit{\(\epsilon_{decay}\)} & 0.9\\
        \hline
    \end{tabular}
    \caption{Hyperparameter used for training}
    \label{tab:hyperparameter}
\end{table}
\subsection{Simulated environment}
\begin{figure}[h]
    \centering
    \includegraphics[width=1\columnwidth]{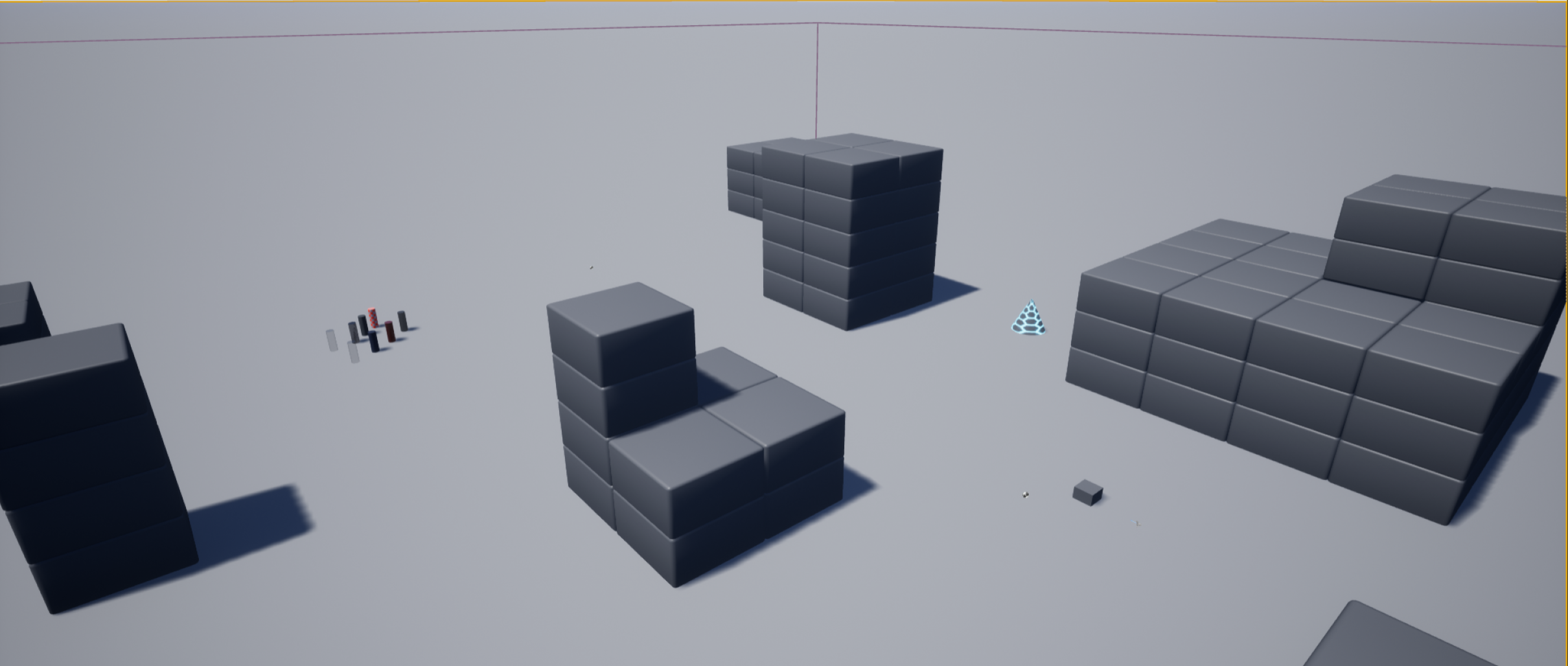}
    \caption{Environment visualization, in clear weather mode}
    \label{fig:env-vis}
\end{figure}

AirSim is an open-source simulation platform developed by Microsoft to facilitate research in autonomous vehicles, including drones and cars. It provides a highly realistic, physics-based environment for training and testing reinforcement learning models, explicitly targeting autonomous navigation, collision avoidance, and object tracking tasks. AirSim is built on the Unreal Engine, allowing for rich visual realism, which is crucial for computer vision applications. It can simulate diverse environmental conditions, such as different weather types and lighting scenarios, to evaluate the robustness of different DRL models. The platform also integrates well with standard machine learning frameworks, enabling training capability for reinforcement learning agents through API. Figure \ref{fig:env-vis} gives a detailed visualization of the operating environment we used to experiment. 

\subsection{Testing scenarios}

\begin{figure*}[ht]
    \centering
    \begin{minipage}[b]{0.2\textwidth}
        \centering
        \includegraphics[width=\textwidth]{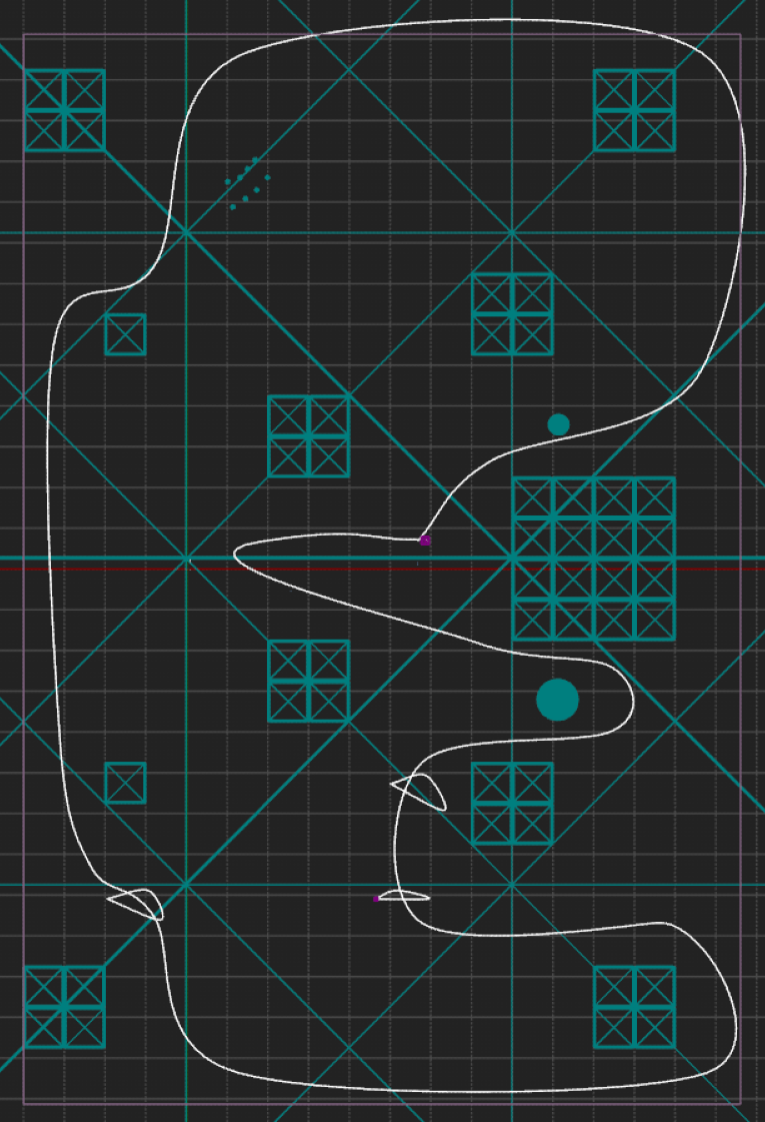}
    \end{minipage}
    \hfill
    \begin{minipage}[b]{0.2\textwidth}
        \centering
        \includegraphics[width=\textwidth]{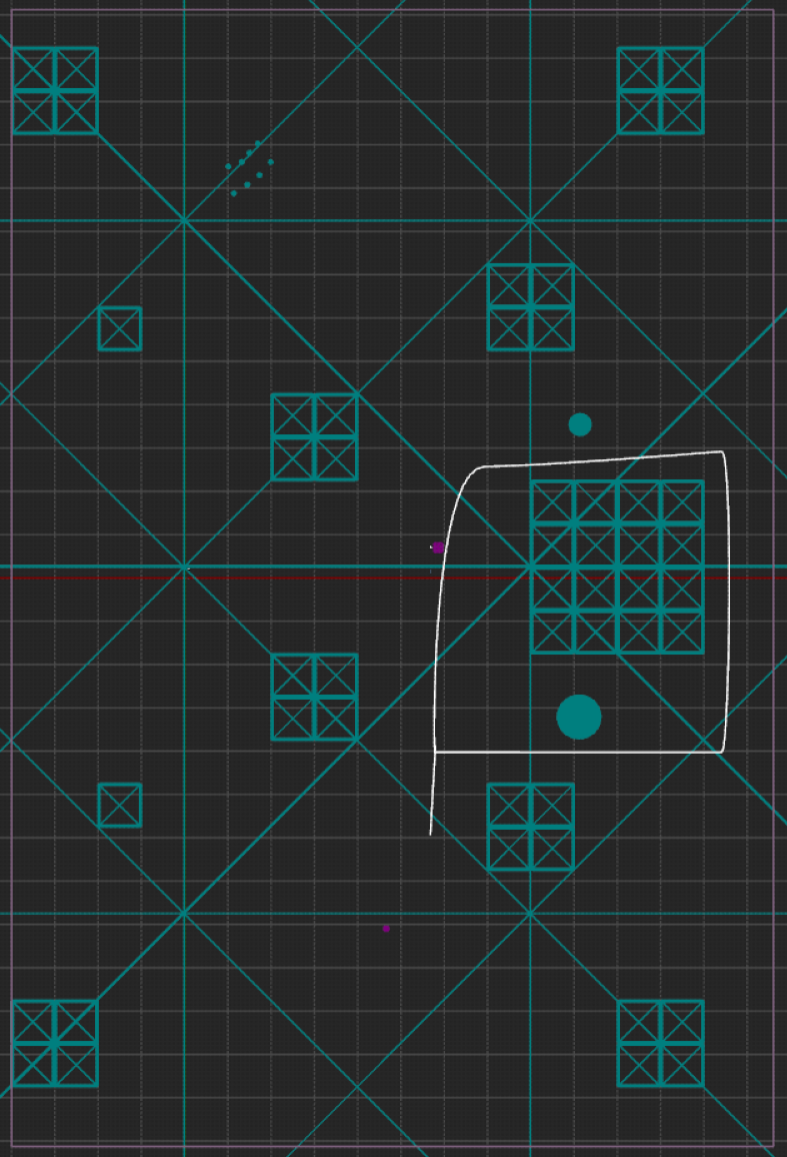}
    \end{minipage}
    \hfill
    \begin{minipage}[b]{0.2\textwidth}
        \centering
        \includegraphics[width=\textwidth]{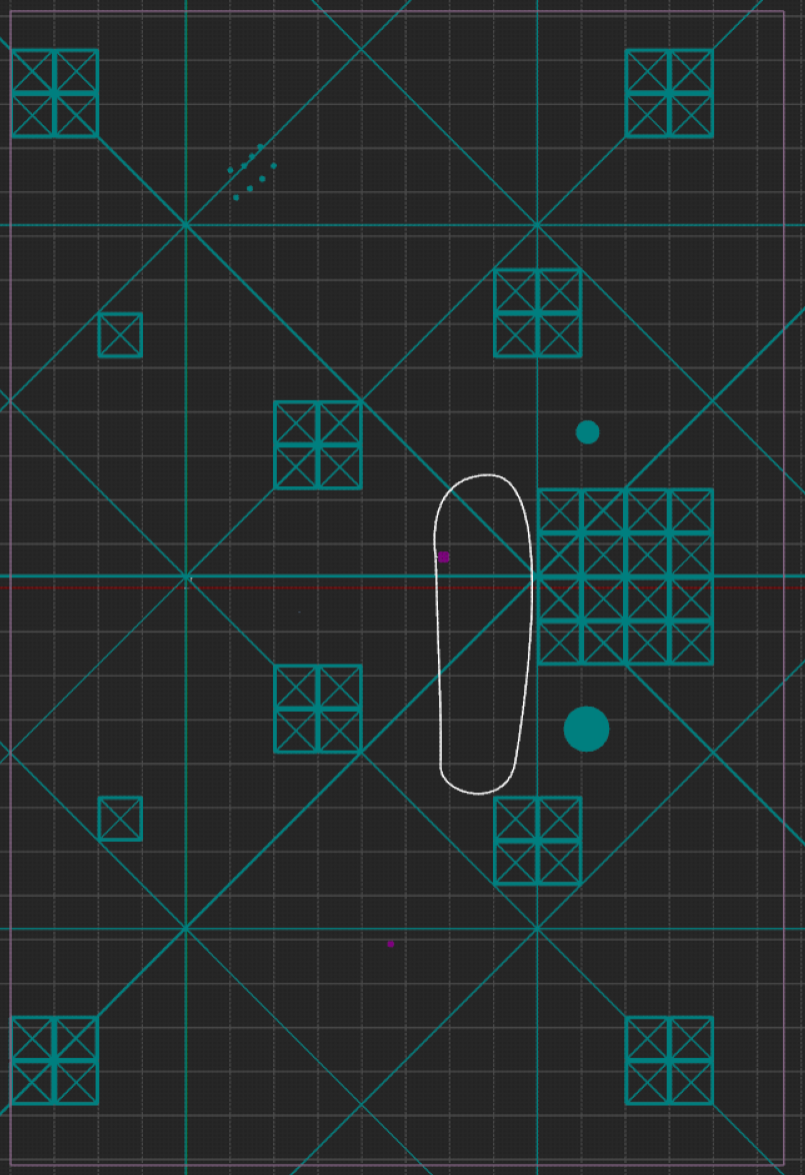}
    \end{minipage}
    \hfill
    \begin{minipage}[b]{0.2\textwidth}
        \centering
        \includegraphics[width=\textwidth]{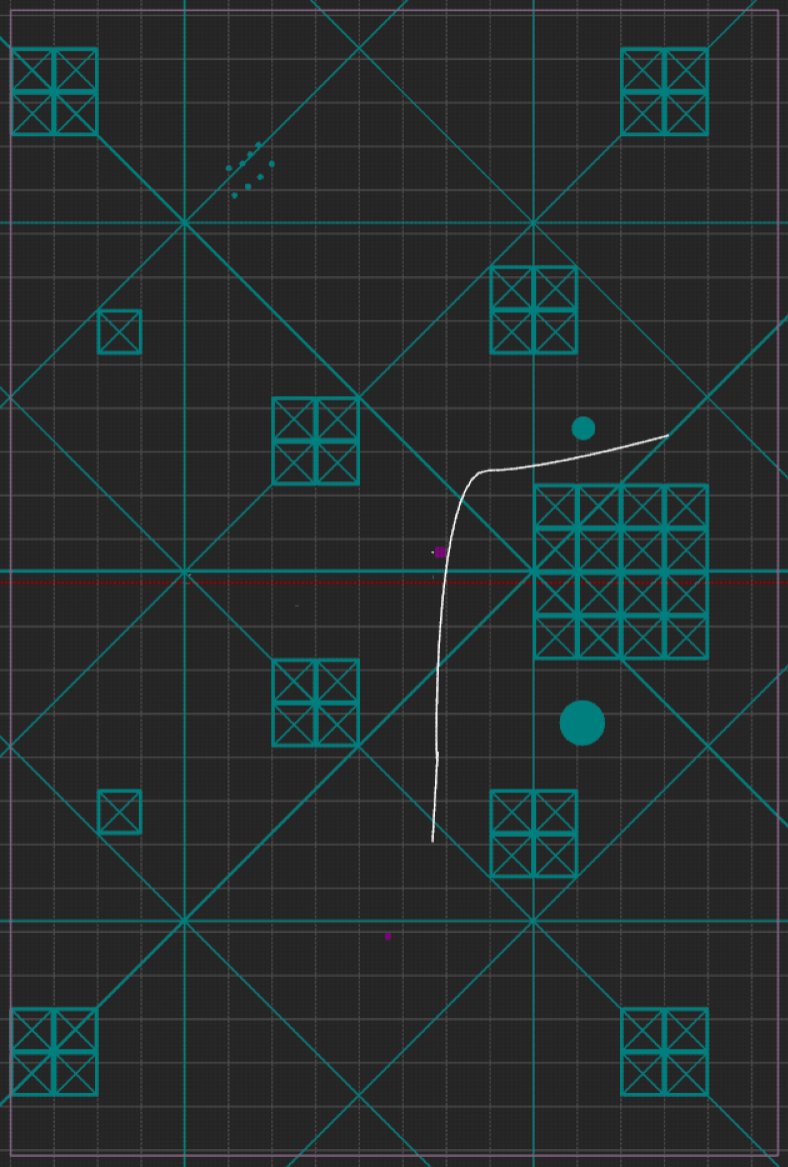}
    \end{minipage}
    \caption{Target path for testing (White line), from left to right: \textit{Complex}, \textit{SharpLoop}, \textit{SimpleLoop}, \textit{SingleTurn}}
    \label{fig:paths}
\end{figure*}

The testing environment consists of four maps designed to evaluate specific aspects of driving performance. These maps offer increasing difficulty and complexity levels, ranging from simple maneuvers to advanced scenarios involving dynamic and static obstacles. Figure \ref{fig:paths} shows the complexity of each map used for training and testing. The descriptions of each map are as follows:

\begin{itemize}
    \item \textit{SingleTurn}: A simple map containing only a 90-degree turn provides a straightforward scenario to evaluate basic turning maneuvers.
    \item \textit{SimpleLoop}: A shape loop designed to test the ability of agents to navigate continuous smooth turns and handle intersections.
    \item \textit{SharpLoop}: A loop around a single obstacle that includes very sharp turns, designed to test the system’s ability to manage tight cornering.
    \item \textit{Complex}: A challenging map that includes turns of all difficulty levels, with dynamic and static obstacles, simulating real-world scenarios with numerous unpredictable elements.
\end{itemize}

We use two metrics to assess the framework's performance: Episode Length (EL) and Cumulative Rewards (CR). EL represents the length of one tracking episode during inference, while CR denotes the sum of global rewards throughout the episode. These metrics have been previously used to comprehensively assess AOT frameworks ~\cite{luo2019end}. Table \ref{tab:map_metrics} shows the maximum episode length and minimum rewards for each map desirably in this experiment. Limiting these two metrics during training also helps maintain meaningful samples for efficient training.
\begin{table}[h!]
    \centering
    \begin{tabular}{|c|c|c|}
        \hline
        \textbf{Map} & \textbf{Maximum EL} & \textbf{Minimum CR}\\
        \hline
        \textit{Complex} & 80 & -150\\
        \hline
        \textit{SharpLoop} & 45 & -150\\
        \hline
        \textit{SimpleLoop} & 25 & -150\\
        \hline
        \textit{SingleTurn} & 15 & -150\\
        \hline
    \end{tabular}
    \caption{Terminating threshold for EL and CR during training}
    \label{tab:map_metrics}
\end{table}

To test the system's performance on an unseen map, we set up the experiment: Train the tracker on a single map for 50 episodes, with given boundaries on EL and CR. Then, we put it to the test on all maps to observe if it can perform well in seen and unseen environments. The final result is the average of the different trials above. With CR, unless the tracker can track the target for the span of the maximum EL, the CR is always be the minimum value by default because of the terminating condition.

Because of the complexity of the task's nature and formulation, it's difficult to compare our proposed method with other ASOT methods without making any changes to the existing solution. Therefore, we implemented the \textbf{SingleAgent} method, which consists of one agent that shares the same agent architecture but without any of our novel adaptations to this task for comparison purposes. The same environment setting is used for this method to compare the performance uniformly. 
\subsection{Results}
Table \ref{tab:metrics} shows the average performance of the tracker in different maps. For trivial maps such as \textit{SimpleLoop} and \textit{SingleTurn}, the CSAOT tracker can achieve high EL, whereas the system's performance isn't decent with those complex maps. We also recognize during the inference run of the CSAOT tracker that it can hardly get back on track if it loses track of the target before. This happened on most maps, excluding the \textit{SingleTurn} map. One thing to be recognized in the experiment is that collision rarely appears throughout the different scenarios, which it only appears in \textit{Complex} map.
\begin{table}[h!]
    \centering
    \begin{tabular}{|c|c|c|}
        \hline
        \textbf{Map} & \textbf{EL} & \textbf{CR} \\
        \hline
        \textit{Complex} & \(25 \pm 4\) & \(-150 \pm 0.00\)\\
        \hline
        \textit{SharpLoop} & \(30\ \pm 3\) & \(-150 \pm 0.00\)\\
        \hline
        \textit{SimpleLoop} & \(23 \pm 1\) & \(-150 \pm 0.00\)\\
        \hline
        \textit{SingleTurn} & \(15 \pm 0\)& \(-25.46 \pm 3.04\)\\
        \hline
    \end{tabular}
    \caption{Average EL and CR of the model during inference on different maps}
    \label{tab:metrics}
\end{table}

Table \ref{tab:comparison} shows the differences in performance between the naive \textbf{SingleAgent} and our novel method \textbf{CSAOT}. With the simplicity of the \textit{SingleTurn} and \textit{SimpleLoop} map, these two methods' performance don't significantly differ. However, the difference is remarkable on \textit{Complex} map, with an increase of approximately 30$\%$ in average EL. This improvement shows that the method did well with the high complexity when the target path is complicated, highlighting the importance of adapting a multi-agent approach to this specific task.
\begin{table}[h!]
    \centering
    \begin{tabular}{|c|l|c|c|}
        \hline
        \textbf{Environment map} & \textbf{Method} & \textbf{AR} & \textbf{EL} \\
        \hline
        \textit{Complex} &\textit{CSAOT} & \textbf{25 $\pm$ 4} & \textbf{-150 $\pm$ 0.00} \\
        & \textit{SingleAgent} & 18 $\pm$ 5 & -150 $\pm$ 0.00  \\
        \hline
        \textit{SharpLoop} &\textit{CSAOT} & \textbf{30 $\pm$ 3} & \textbf{-150 $\pm$ 0.00} \\
        & \textit{SingleAgent} & 26 $\pm$ 2 & -150 $\pm$ 0.00  \\
        \hline
        \textit{SimpleLoop} &\textit{CSAOT} & \textbf{23 $\pm$ 1} & \textbf{-150 $\pm$ 0.00} \\
        & \textit{SingleAgent} & 21 $\pm$ 2 & -150 $\pm$ 0.00 \\
        \hline
        \textit{SingleTurn} & \textit{CSAOT} & \textbf{15 $\pm$ 0} & \textbf{-25.46 $\pm$ 3.04} \\
        & \textit{SingleAgent} & \textbf{15 $\pm$ 0} & -31.37 $\pm$ 2.78 \\
        \hline
    \end{tabular}
    \caption{Performance comparison between \textbf{SingleAgent} approach and \textbf{CSAOT}. Values written in bold denote best performance}
    \label{tab:comparison}
\end{table}

\begin{figure}[h]
    \centering
    \includegraphics[width=1\columnwidth]{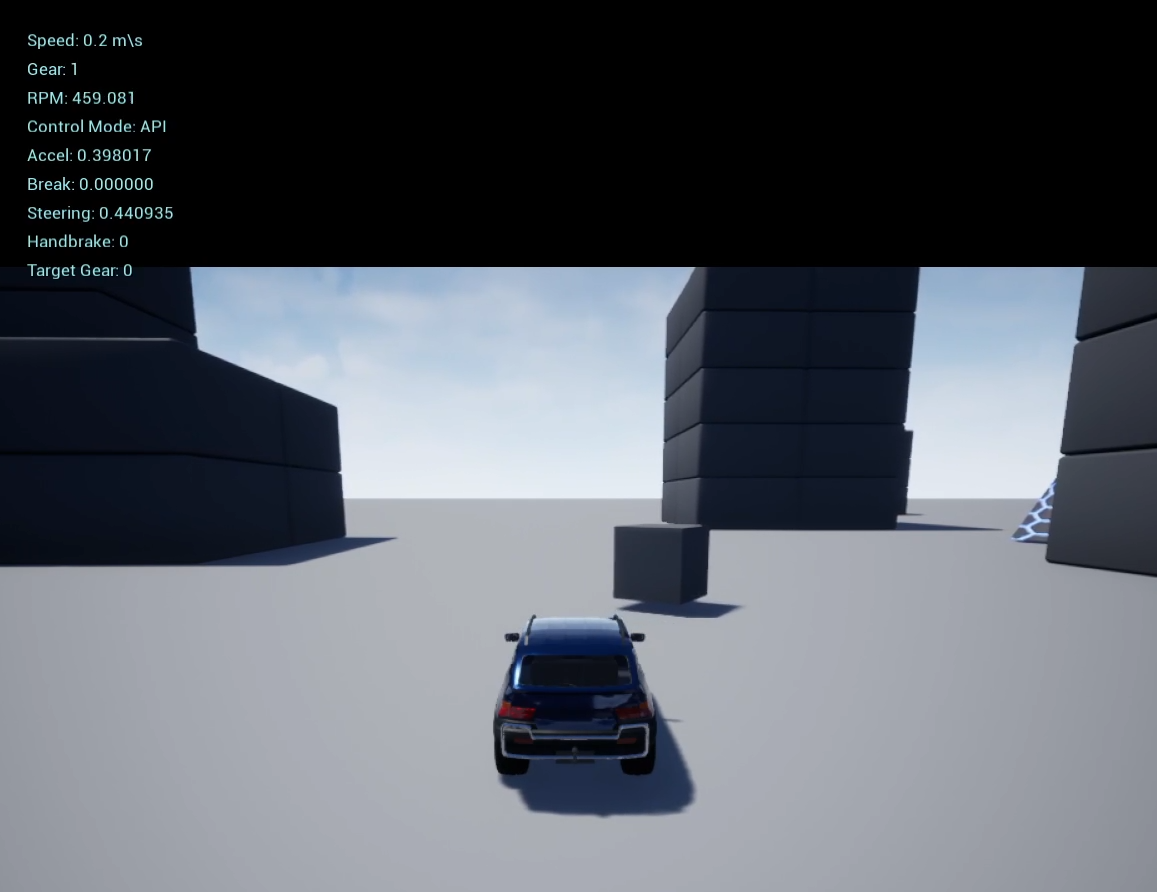}
    \caption{Sample decision in \textit{Complex} map. Steer and acceleration values are 0.398 and 0.44, respectively}
    \label{fig:sample-decision}
\end{figure}

To see the system's actual performance in inference time, we investigated the system's decisions at runtime. For the trivial maps, it tracks the target closely while fulfilling the desired behavior that we have defined through the reward function. On the other hand, in \textit{Complex} map, it can only track within the first 3 turns, then loses track of the target due to inappropriate turning. Figure \ref{fig:sample-decision} shows a sample moment from one episode under \textit{Complex} map. As can be seen, the actions from those first steps are accurate, showing that the system is learning in the proper direction. We provide a sample episode footage as supplementary material.

\section{Discussion}
\subsection{System architecture}
Although the system's architecture has improved performance compared to its single-agent counterpart, we believe there are different options for alternating and experimenting with the modules. Recently, State Space Model and its variants such as S4 ~\cite{gu2022efficientlymodelinglongsequences} or Mamba ~\cite{gu2024mambalineartimesequencemodeling}, which have great potential to perform better than our usage of LSTM for memorial subtask. Hence, we consider this part of our future work on this framework.
\subsection{Learning strategy}
\begin{figure}[h]
    \centering
    \includegraphics[width=1\columnwidth]{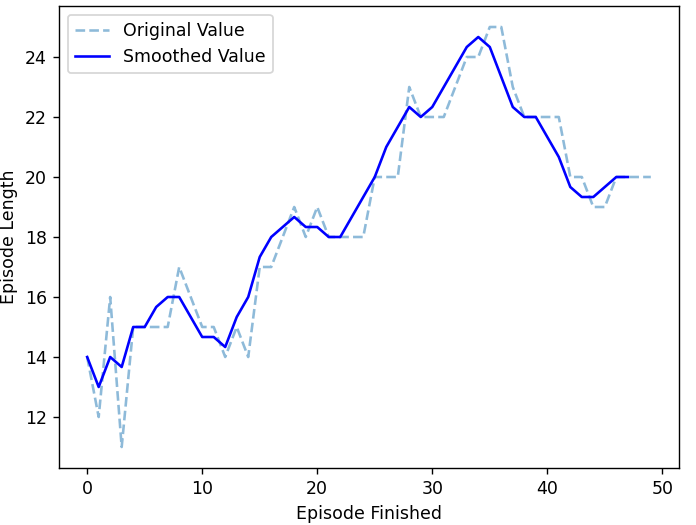}
    \caption{EL during training with CSAOT in \textit{Complex} map}
    \label{fig:el-complex}
\end{figure}

PPO is shown to be sample-efficient in previous work ~\cite{schulman2017proximalpolicyoptimizationalgorithms}, and it's been shown clearly through our experiment. However, the rapid convergence of this method increases the probability of falling into local minima during training. This can be observed through the learning process of the CSAOT tracker in \textit{Complex} map, which figure \ref{fig:el-complex} provides more insights about this convergence. This seems to be a common phenomenon for actor-critic-based methods. To avoid this, methods to trigger further exploration or adding an early stopping mechanism could help. In our implementation, we make use of \(\epsilon\)-greedy to tackle the problem, but it doesn't seem to work well. More advanced methods, such as the Adversarially Guided Actor-Critic (AGAC) ~\cite{fletberliac2021adversariallyguidedactorcritic} mechanism, could help improve the overall learning performance of the system.
\subsection{Gating mechanism for MoP}
Our proposed policy network idea combines policies to increase the policy's robustness and adaptability in different scenarios. However, we haven't considered the effect of the gating mechanism within the MoP architecture. We hypothesize that this can be an essential factor affecting the framework's performance in such a complex system setting. Therefore, further investigation to choose the proper mechanism can improve the learning process's accuracy.
\subsection{Intrinsic reward for individual agent}
We introduced multiple task-based component rewards to improve the system's convergence speed, and compared to the \textbf{SingleAgent} implementation, the rewards help enhance the performance in a high complexity environment. One way to increase the effect of those rewards is to add some intrinsic reward in addition to the extrinsic reward. Intrinsic reward has been proven helpful in cooperative MADRL systems ~\cite{9892092}, and we will investigate further into this field.

\section{Conclusion}

In this work, we proposed CSAOT, a novel active tracker for SOT via MADRL. Unlike previous work on ASOT, the proposed tracker comprises a multi-agent system that sits on a single vehicle, efficiently utilizing the information extracted from the environment. Furthermore, we proposed MoP as an alternative to the traditional policy network of an MLP block. Combining those factors, the system is proven to perform well in a multi-dimensional continuous action space and generalizes to unseen environments. We believe that this framework can be applied to real-world scenarios.



\bibliographystyle{ACM-Reference-Format} 
\bibliography{sample}


\end{document}